\documentclass[letterpaper]{article} 
\usepackage{aaai25}  
\usepackage{times}  
\usepackage{helvet}  
\usepackage{courier}  
\usepackage[hyphens]{url}  
\usepackage{graphicx} 
\usepackage{natbib}  
\usepackage{caption} 
\usepackage{algorithm}
\usepackage{algorithmic}
\usepackage{times}
\usepackage{epsfig}
\usepackage{graphicx}
\usepackage{amsmath}
\usepackage{amssymb}
\usepackage{booktabs}
\usepackage{multirow}
\usepackage{threeparttable}
\usepackage{xcolor}
\usepackage{newfloat}
\usepackage{listings}
\DeclareCaptionStyle{ruled}{labelfont=normalfont,labelsep=colon,strut=off} 
\floatstyle{ruled}
\newfloat{listing}{tb}{lst}{}
\floatname{listing}{Listing}
\title{Exploring More from Multiple Gait Modalities for Human Identification}
\author{
    Dongyang Jin\textsuperscript{\rm 1}\equalcontrib, 
    Chao Fan\textsuperscript{\rm 1}\equalcontrib,
    Weihua Chen\textsuperscript{\rm 2},
    Shiqi Yu\textsuperscript{\rm 1}\thanks{Corresponding author.}
}
\affiliations{
    \textsuperscript{\rm 1}Department of Computer Science and Engineering, Southern University of Science and Technology, China\\
    \textsuperscript{\rm 2}Alibaba Group \\
    \{12332451, 12131100\}@mail.sustech.edu.cn, kugang.cwh@alibaba-inc.com, yusq@sustech.edu.cn
}

\usepackage{bibentry}

\begin{document}

\maketitle

\begin{abstract}
The gait, as a kind of soft biometric characteristic, can reflect the distinct walking patterns of individuals at a distance, exhibiting a promising technique for unrestrained human identification. 
With largely excluding gait-unrelated cues hidden in RGB videos, the silhouette and skeleton, though visually compact, have acted as two of the most prevailing gait modalities for a long time. 
Recently, several attempts have been made to introduce more informative data forms like human parsing and optical flow images to capture gait characteristics, along with multi-branch architectures. 
However, due to the inconsistency within model designs and experiment settings, we argue that a comprehensive and fair comparative study among these popular gait modalities, involving the representational capacity and fusion strategy exploration, is still lacking. 
From the perspectives of fine vs. coarse-grained shape and whole vs. pixel-wise motion modeling, this work presents an in-depth investigation of three popular gait representations, i.e., silhouette, human parsing, and optical flow, with various fusion evaluations, and experimentally exposes their similarities and differences. Based on the obtained insights, we further develop a C$^2$Fusion strategy, consequently building our new framework MultiGait++. C$^2$Fusion preserves commonalities while highlighting differences to enrich the learning of gait features.
To verify our findings and conclusions, extensive experiments on Gait3D, GREW, CCPG, and SUSTech1K are conducted. The code is available at \url{https://github.com/ShiqiYu/OpenGait}. 
\end{abstract}

\section{Introduction}
\label{sec:intro}
\begin{figure}[t]
\centering
\includegraphics[width=0.9\columnwidth]{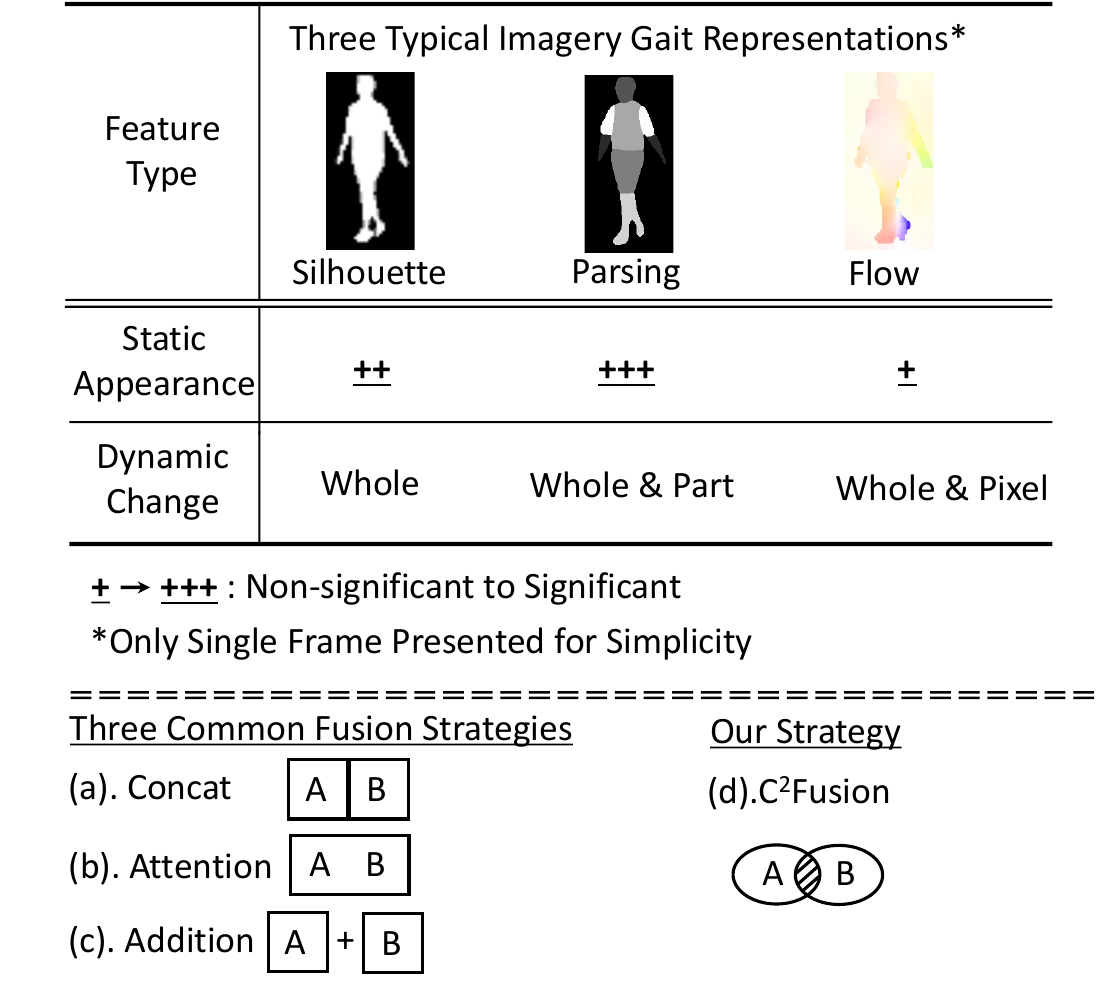}
\caption{Top: comparing three typical gait modalities, i.e., the binary silhouette, body parsing, and optical flow images. Bottom: the comparison between three common fusion strategies with our C$^2$Fusion. }
\label{fig:intro}
\end{figure}

The pedestrian gait presented in walking videos typically involves the visual characteristics of body shape and limb movements. 
This unique application, known as gait recognition, distinguishes itself from other biometrics techniques such as face, fingerprint, and iris recognition, by offering the flexibility of non-intrusive and long-distance usage without necessitating the subject's cooperation. 
Furthermore, gait is inherently hard to disguise and conceal. 
These attributes render gait recognition particularly well-suited for unconstrained security applications such as suspect tracking and retrieval~\cite{nixon2006automatic}. 

To mitigate the influence of irrelevant visual cues such as texture and background, gait recognition methods often rely on derived gait representations extracted from RGB videos rather than the videos themselves as inputs.
The widely used ones include binary silhouettes~\cite{Chao2019,fan2020gaitpart,gaitgl,shen2023gait,wang2024qagait,fan2023exploring, fan2023learning}, skeleton coordinates~\cite{liao2017pose,teepe2021gaitgraph,fan2023skeletongait}, SMPL model~\cite{li2020end,zheng2022gait3d}, body parsing~\cite{zheng2023parsing}, and optical flow images~\cite{castro2024attengait,feng2023fusion}.  
Among these, the silhouette is favored for its clarity in color and richness in shape, making it the most popular choice for gait modeling.
Recent studies have pointed out the limitations in silhouette images, citing a lack of fine-grained part-level shape~\cite{zheng2023parsing} and explicit body structural characteristics~\cite{peng2024learning}, and introduce extra body parsing images and human joints coordinate, often driven by multi-branch networks, to reach the new state-of-the-art recognition performance. 
In this paper, we agree that the multimodal approach represents a pivotal direction for advancing gait recognition research. 

Despite ongoing efforts, a fair and comprehensive comparative study among some representative gait modalities remains necessary, due to the usual misalignments in model and experimental settings.
To advance this understanding, this work investigates three typical gait modalities: silhouette, human parsing, and optical flow images, as depicted in Figure~\ref{fig:intro}.
These modalities, each with distinct physical meanings, emphasize different aspects of gait: whole-body shapes, body parts, and pixel-level motion dynamics, respectively, offering varied levels of feature granularity. 
Among them, optical flow images are particularly notable for capturing pixel-level motion in each frame, setting them apart from silhouettes and human parsing images that primarily focus on body shapes.
While these modalities have their differences, they also share similarities. 
For instance, the overall contours of the human body can be represented by both binary silhouettes and human parsing images. 
All three modalities convey dynamic changes in human limbs.
Leveraging three common fusion strategies depicted in Figure~\ref{fig:intro} (a-c), this work introduces the MultiGait series, i.e., a set of uni- and multimodal baselines, that provide a comprehensive study on gait modality fusion.
This approach allows for a fair and thorough examination of the efforts of each modality for multimodal gait modeling.

Building on the exploration of similarities and differences among silhouette, body parsing, and optical flow images, this work goes a step further by proposing a novel gait modality fusion strategy termed C$^2$Fusion, leading to the development of a new multimodal method named MultiGait++. 
As illustrated in Figure~\ref{fig:intro}(d), the core idea of C$^2$Fusion is to extract shared characteristics across different modalities while simultaneously encouraging each modality to emphasize its unique attributes beyond these commonalities.
This design forces MultiGait++ to fully explore the diverse features offered by given modalities, thereby enhancing its discriminative capacity. 
Experiments show that MultiGait++ achieves a new SoTA on the Gait3D~\cite{zheng2022gait3d}, GREW~\cite{zhu2021gait}, CCPG~\cite{li2023depth} and SUSTech1K~\cite{Shen_2023_CVPR} datasets.

Overall, this work contributes to the field in two-fold:
\begin{itemize}
  \item \textbf{An In-Depth Gait Modality Fusion Study under Fair Conditions:} MultiGait marks one of the first comparative studies on uni- vs. multimodal gait recognition under fair conditions. Through extensive experiments, we comprehensively examine the specific contributions and limitations of modalities such as silhouette, human parsing, and optical flow images for gait description. 
  \item \textbf{A Novel Multimodal Gait Recognition Method:} 
  We introduce MultiGait++, featured by its core component, C$^2$Fusion.
  This approach maximizes the extraction of diverse features from given modalities, thereby enhancing the overall representation of gait patterns.
\end{itemize}

\section{Related Work}

\subsection{Gait Modalities}
Gait modalities are captured by a range of sensors, including traditional RGB cameras as well as emerging technologies like event cameras~\cite{9337225}, LiDAR~\cite{Shen_2023_CVPR,wang2023pointgait,Wang2024Cross,guo2025camera}, and fisheye cameras~\cite{xu2023gait}. 
Among these, RGB-based cameras continue to dominate gait recognition due to their cost-effectiveness and seamless integration with existing CCTV systems. 
Consequently, the following literature review focuses on gait modalities extracted from RGB images, which include binary silhouettes, optical flow images, human parsing, and 2D/3D poses. Each modality offers distinct advantages, yet they all aim to minimize the influence of gait-irrelevant factors such as clothing color, texture, and background. We categorize gait recognition methods into two groups based on the number of modalities used: unimodal and multimodal.

\subsection{Unimodal Gait Recognition Methods}
This kind of method usually extracts gait features from the sequence of binary silhouettes, 2D/3D coordinates, human parsing, or optical flow images. 

With the rapid advancement of deep learning, silhouette-based methods primarily focus on extracting both spatial and temporal features of gait. For instance, GaitSet~\cite{Chao2019} treats the gait sequence as a set and uses a maximum function to compress frame-level spatial features. 
GaitPart~\cite{fan2020gaitpart} pays more attention to spatial and temporal details, showcasing their significance. 
GaitGL~\cite{lin2022gaitgl} introduces a GLConv block to capture global and local spatial features simultaneously. 
The latest OpenGait~\cite{fan2024opengait} offers a comprehensive exploration of deep model design for outdoor gait recognition, achieving strong performance across various datasets.

In pose-based methods, PoseGait~\cite{liao2020model} combines 3D skeleton data with hand-crafted features to address the challenges posed by changes in viewpoint and clothing. 
GaitGraph~\cite{teepe2021gaitgraph} uses a graph convolution network to learn gait modalities based on 2D skeletons.
SkeletonGait~\cite{fan2023skeletongait} proposes an image-like skeleton modality to enhance gait feature learning, offering novel insights into the role of body structure characteristics.

Human parsing is visually similar to silhouettes that provide fine-grained body part annotations. 
Recent studies on parsing-based gait recognition~\cite{zou2024cross} typically focus on modeling both the entire body and individual body parts.
For example, GaitParsing~\cite{wang2023gaitparsing} addresses the self-occlusion problem through human parsing. ParsingGait~\cite{zheng2023parsing} extracts holistic body features while incorporating a GCN branch to capture structural relationships among various body parts.

Individuals have unique movements and speeds, making movement behavior a crucial aspect of human identity recognition.
While silhouettes and human parsing focus on body shape, optical flow is gaining attention for its representation of instantaneous motion.
Ye~\cite{ye2023gait} and Xu~\cite{xu2023attention} proposed that the motion information in optical flow can better match individuals with similar body shapes. AttenGait~\cite{castro2024attengait} demonstrates that the modality information in optical flow is richer than that in silhouettes, showing great potential for gait recognition.

\subsection{Multimodal Gait Recognition Methods}
An increasing number of methods are now focused on extracting rich features from multiple gait modalities.
For instance, SMPLGait~\cite{zheng2022gait} utilizes the 3D SMPL model to refine the learning from gait silhouettes.
BiFusion~\cite{peng2024learning} integrates skeletons and silhouettes to capture the comprehensive spatiotemporal characteristics of human gait. 
Feng~\cite{feng2023fusion} explored the complementary nature of movement and shape information by combining silhouettes and optical flow images. 
XGait~\cite{zheng2024takes} proposes a novel cross-granularity alignment method to unleash the power of gait modalities of different granularity. 
Additionally, SkeletonGait++~\cite{fan2023skeletongait} combines silhouettes and the proposed skeleton maps in a frame-by-frame manner, fully leveraging the strengths of CNNs. 
In this work, we propose a new idea that extracts common characteristics across different modalities while simultaneously encouraging each modality to express its unique attributes, thereby enhancing the learning of rich multimodal gait features.

\section{Method}
\label{sec:method}
\subsection{Silhouette, Parsing, and Optical Flow for Gait}
\label{sec:explore}

Here we study three mainstream gait modalities: \textbf{silhouette, human parsing, and optical flow}. 
In previous research, silhouette has been the primary focus, while human parsing presented an emerging topic showing significant potential. 
Although optical flow has been less shiny in recent leading venues for gait recognition, we include it due to its fine-grained nature for motion description.

\begin{figure}[t]
\centering
\includegraphics[width=0.95\columnwidth]{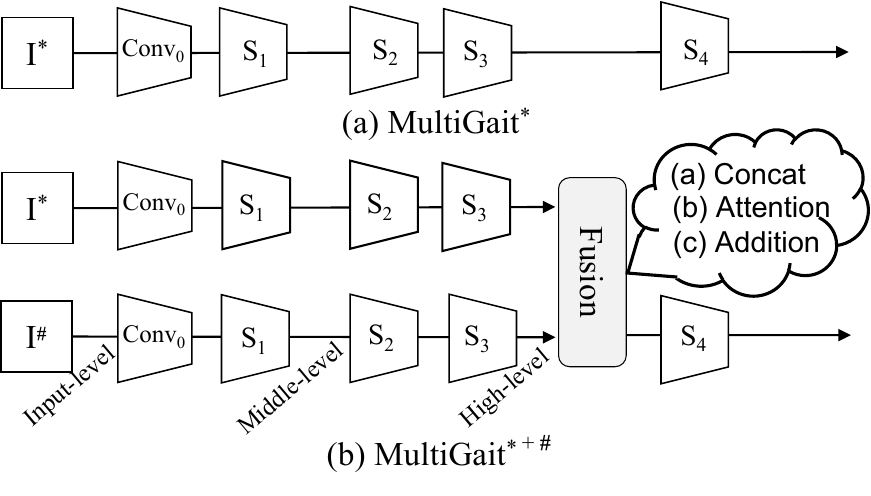}
\caption{
The architecture of the MultiGait series. 
Here the symbols * and \# can be instantiated with any of the employed gait modalities, such as the silhouette, human parsing, and optical flow, in theory. 
}
\label{fig:MultiGait}
\end{figure}

To ensure an intuitive and fair investigation, we adopt a uniform framework to model these three modalities and their combinations. 
Structurally, we utilize the architecture of DeepGaitV2~\cite{fan2024opengait} thanks to its straightforward design and strong performance. 
As a result, three unimodal baselines, \textit{i.e.}, MultiGait$^s$ (totally identical to DeepGaitV2), 
MultiGait$^p$, MultiGait$^f$, as well as two multimodal ones, \textit{i.e.}, MultiGait$^{s+p}$ and MultiGait$^{s+f}$, are shown in Figure.~\ref{fig:MultiGait}, where $s, p$ and $f$ denote the input of silhouette, human parsing, and optical flow images, respectively. 

These baselines share identical blocks, with the multimodal ones employing additional naive multi-branch structures. 
Additionally, we introduce MultiGait$^{2s}$, which doubles the channels of MultiGait$^{s}$, to eliminate the effects of increased parameters brought by multi-branch designs.
To enrich the research scope, we consider various feature fusion locations (input, middle, and high level) and mechanisms (element-wise addition, channel-wise concatenation, and cross-attention fusion\footnote{
Here we utilize the popular cross-attention mechanism from SkeletonGait++~\cite{fan2023skeletongait}.
}), as shown in Figure.~\ref{fig:MultiGait} (b).
For now, we defer the implementation details to focus on key insights that advance multimodal gait recognition and inspire the design principles behind our MultiGait++.

\begin{table*}[t]
\centering
\begin{tabular}{c|c|c|cccccccc|cc}
\toprule
\multirow{2}{*}{Method}    & \multirow{2}{*}{\begin{tabular}[c]{@{}c@{}}Fusion\\ Location\end{tabular}} & \multirow{2}{*}{\begin{tabular}[c]{@{}c@{}}Fusion\\ Mechanism\end{tabular}} & \multicolumn{8}{c|}{Probe Sequence (Rank-1 acc)}        & \multicolumn{2}{c}{Overall} \\
                           &                        &                            & NM   & BG    & CL   & CR   & UB   & UN   & OC   & NT   & R-1           & R-5           \\ \midrule
\multirow{1}{*}{MultiGait$^{s}$} & \multicolumn{2}{c|}{N/A}                      & 86.5 & 82.8  & 49.2 & 80.4 & 83.3 & 81.9 & 86.0 & 28.0 & 80.9         & 91.9         \\ 
\multirow{1}{*}{MultiGait$^{p}$} & \multicolumn{2}{c|}{N/A}                      & 84.7 & 77.3  & 29.4 & 78.2 & 75.8 & 80.2 & 87.9 & 43.3 & 77.3         & 91.3         \\ 
\multirow{1}{*}{MultiGait$^{f}$} & \multicolumn{2}{c|}{N/A}                      & 84.4 & 82.8  & 54.9 & 79.3 & 82.4 & 79.8 & 89.7 & 35.0 & 80.5         & 92.5          \\ \midrule
\multirow{1}{*}{MultiGait$^{2s}$} & \multicolumn{2}{c|}{N/A}                    & 87.6 & 83.9  & 50.1 & 81.3 & 85.1 & 83.6 & 86.7 & 29.1 & 81.9         & 92.3          \\ \midrule
\multirow{7}{*}{MultiGait$^{s+p}$} & \multirow{1}{*}{Input} & Concatenation      & 89.6 & 87.1 & 45.1 & 85.4 & 87.4 & 86.7 & 90.3 & 43.2 & 85.1         & 94.9         \\ \cmidrule{2-13} 
                           & \multirow{3}{*}{Middle}   & Concatenation              & 90.6 & 86.5  & 44.6 & 85.7 & 86.7 & 86.6 & 91.2 & 44.5 & 85.2         & 94.9         \\
                           &                        & Attention                  & 90.2 & 86.7  & 43.5 & 85.6 & 87.0 & 87.1 & 91.5 & 43.0 & 85.1         & 94.8         \\
                           &                        & Addition                   & 89.5 & 87.0  & 41.5 & 85.7 & 87.2 & 86.6 & 91.4 & 45.1 & 85.2         & 94.9         \\ \cmidrule{2-13} 
                           & \multirow{3}{*}{High}  & Concatenation              & 90.9 & 86.7  & 42.4 & 85.8 & 87.4 & 87.3 & 92.3 & 44.5 & 85.4         & 94.9         \\
                           &                        & Attention                  & 90.4 & 85.9  & 38.4 & 84.9 & 86.8 & 86.4 & 91.3 & 43.6 & 84.5         & 94.5         \\
                           &                        & Addition                   & 91.1 & 87.3  & 43.5 & 86.2 & 88.0 & 87.4 & 92.3 & 45.5 & 85.8         & 95.0         \\ \midrule
\multirow{7}{*}{MultiGait$^{s+f}$} & \multirow{1}{*}{Input} & Concatenation      & 86.3 & 83.2  & 51.5 & 80.8 & 84.1 & 82.5 & 86.5 & 27.8 & 81.3         & 92.1         \\ \cmidrule{2-13} 
                           & \multirow{3}{*}{Middle}   & Concatenation              & 87.0 & 83.7  & 51.8 & 81.2 & 84.1 & 82.3 & 87.0 & 29.3 & 81.7         & 92.4         \\ 
                           &                        & Attention                  & 87.4 & 84.2  & 52.0 & 81.5 & 84.9 & 82.6 & 87.5 & 28.6 & 82.1         & 92.6         \\
                           &                        & Addition                   & 87.4 & 84.1  & 52.8 & 81.6 & 85.2 & 83.3 & 87.7 & 28.4 & 82.3         & 92.5         \\ \cmidrule{2-13} 
                           & \multirow{3}{*}{High}  & Concatenation              & 88.8 & 86.1  & 55.2 & 83.2 & 87.0 & 84.6 & 88.5 & 31.0 & 83.9         & 93.4         \\
                           &                        & Attention                  & 89.1 & 85.4  & 54.1 & 82.7 & 86.3 & 84.7 & 88.9 & 31.2 & 83.5         & 93.2         \\
                           &                        & Addition                   & 88.6 & 85.5  & 55.2 & 83.1 & 86.8 & 84.2 & 88.6 & 32.0 & 83.7         & 93.5         \\ \bottomrule
\end{tabular}
\caption{Recognition results of MultiGait on SUSTech1K.}
\label{tab:fusion}
\end{table*}

\begin{figure*}[t]
  \centering
   \includegraphics[width=0.9\linewidth]{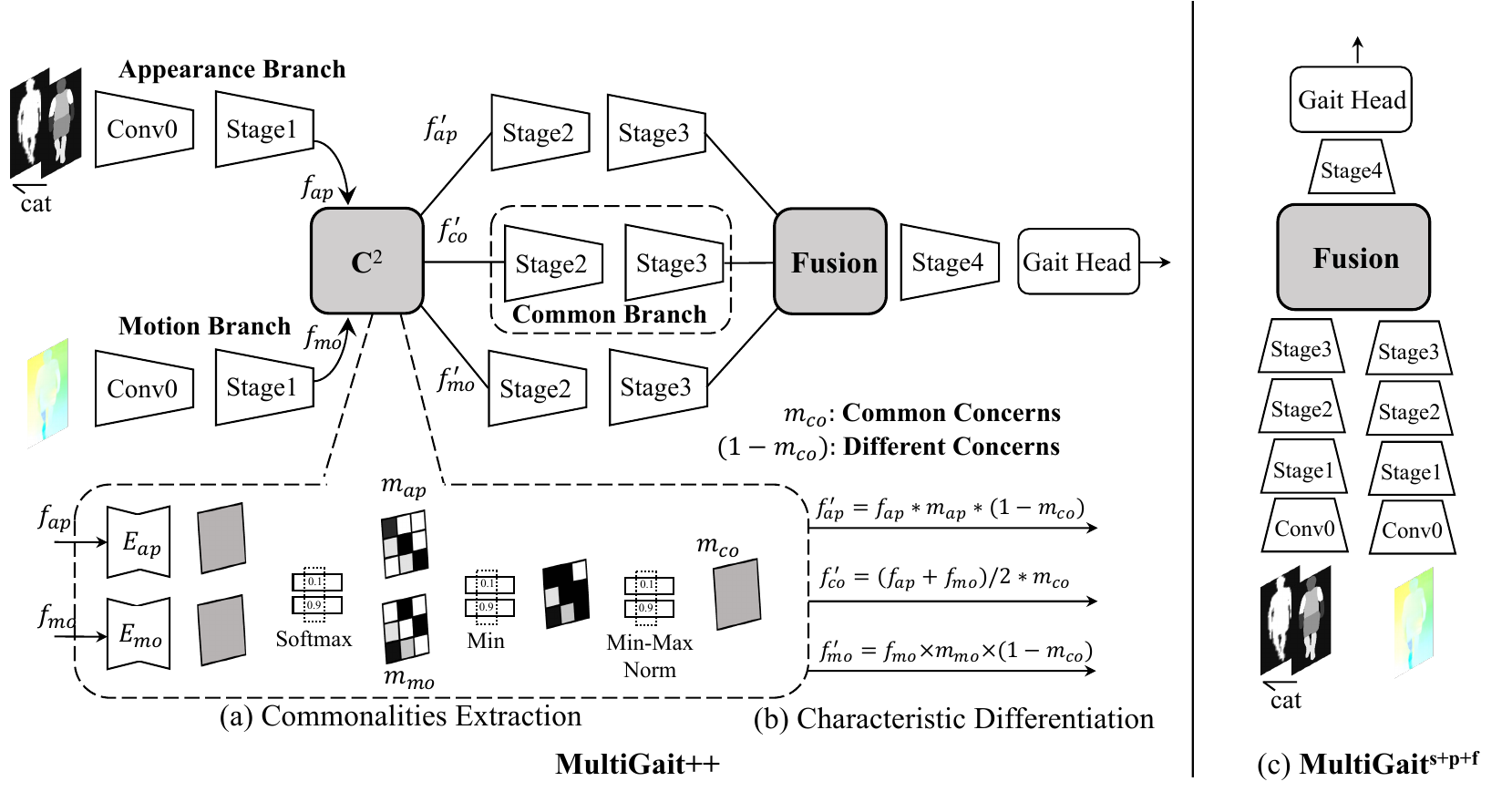}
   \caption{Left: Our pipeline of MultiGait++. Right: The architecture of MultiGait$^{s+p+f}$ }
   \label{fig:MultiGait++}
\end{figure*}

\noindent\textbf{Human Parsing:}
Intuitively, human parsing images offer more detailed cues of body part labels, shapes, and structures compared to binary silhouettes.
This should ideally allow for a more fine-grained interpretation of human gait. 
However, as shown in Table~\ref{tab:fusion}, MultiGait$^p$ does not meet these expectations.
We attribute this discrepancy to the inherently complex nature of human parsing extraction, which, while providing additional characteristics, can also introduce potential noise due to its fine-grained segmentation granularity compared to silhouettes.
This also explains why existing methods tend to combine the silhouette and human parsing rather than replacing the former with the latter. 

\noindent\textbf{Optical Flow:}
Optical flow images excel at capturing pixel-level motion but struggle to represent body shapes (the major advantage of gait silhouettes). 
Surprisingly, Table~\ref{tab:fusion} shows that MultiGait$^f$ achieves highly competitive performance compared with the silhouette-based MultiGait$^s$.
This highlights the untapped potential of optical flow in enhancing gait description. 

\noindent\textbf{Silhouette+Human Parsing:} 
Table~\ref{tab:fusion} indicates that directly combining silhouettes and human parsing images, \textit{i.e.}, MultiGait$^{s+p}$, can enhance the performance significantly.
Moreover, we observe that these gains are nearly uniform over various fusion locations and mechanisms. 
This suggests a high degree of homogeneity in how these two modalities describe human gait, despite their differences in detail. 
Therefore, this work plans to merge them at the input level, saving the model's complexity and computational demands while preserving SoTA performance.

\noindent\textbf{Silhouette+Optical Flow:}
We observe that integrating silhouettes and optical flow images, \textit{i.e.}, MultiGait$^{s+f}$, at a higher level will result in more performance improvements. 
This suggests that silhouettes, which represent body shape, and optical flow, which capture pixel-level motion, complement each other but are not homogeneous in data forms. 
By fusing them at a high level, the combined utility is enhanced. 

Based on the above comprehensive study, we present solid evidence showing that: 
\begin{itemize}
  \item The silhouette provides accurate whole-body shapes, while human parsing adds detailed body part features. Thanks to homogeneity in data forms, they can be effectively fused at the input level. On the other hand, optical flow presents distinct pixel-wise motion characteristics and should be appended at a higher level to capture its unique contributions.
  \item 
  Exploring the shared discriminative features across these three gait modalities, while also identifying and leveraging their unique characteristics, is essential for enriching the learning of multimodal gait representations.
\end{itemize}

\subsection{MultiGait++}
\noindent\textbf{Motivation:} 
Silhouettes, human parsing, and optical flow images are all image-based gait modalities that, despite their differences, naturally share a substantial amount of common features.
Recognizing this point, we propose to encourage each branch to highlight its unique discriminative characteristics beyond these commonalities, thereby enriching the features for multimodal gait description. 

\noindent \textbf{Overall:}
In line with practices established by MultiGait series, we develop a new multimodal method named MultiGait++. 
Specifically, MultiGait++ processes each frame equally and fuses them in a frame-by-frame manner, thus the following formulation considers a single frame for brevity.

As shown in Figure~\ref{fig:MultiGait++}, 
MultiGait++ employs a three-stream architecture.
The appearance branch takes the concatenation of silhouette and human parsing image as input, while the parallel motion branch processes the optical flow image. 
It is important to note that both branches capture different granularities of body shape and dynamic features, with the terms `appearance' and `motion' used here only for clarification in presentation.

The first component of C$^2$Fusion, i.e., the C$^2$ module, then extracts shared features of the above two branches to form an additional common branch.
Meanwhile, the C$^2$ module also refines the output features of these two branches to emphasize their differences. 

After Stage2 and 3, another component of C$^2$Fusion, a straightforward concatenation fusion operation, then aggregates the features from all three branches. 

The final Stage4 and gait head project the rich features extracted from multiple gait modalities into the identity metric space. 
Following the general practices summarized by OpenGait~\cite{fan2022opengait}, the gait head comprises several widely used components, including temporal pooling, horizontal pooling, separate fully connected layers, and BNNecks~\cite{luo2019bag}. 
The overall training process is driven by the triplet and softmax losses. 

In the following section, we focus on the key component of C$^2$Fusion, specifically the C$^2$ module.

\noindent\textbf{C$^2$Fusion:}
Through the Conv0 and Stage1, the appearance and motion branch respectively output feature $f_{ap}$ and $f_{mo}$ with a shape of $C\times H\times W$ (channel$\times$ height $\times$ width).

As shown in Figure~\ref{fig:MultiGait++} (a), 
the C$^2$ module initiates a global understanding through a cross-attention operation: 
\begin{equation}
    m_{ap}, m_{mo} = \text{Softmax}(E_{ap}(f_{ap}), E_{mo}(f_{mo})), 
\end{equation}
where $E_{ap}$ and $E_{mo}$ represent two simple squeeze-and-excitation networks with a squeeze rate of 16.
With the help of an element-wise softmax function, these networks project the $f_{ap}$ and $f_{mo}$ to an identical-size attention map $m_{ap}$ and $m_{mo}$, respectively. 

Intuitively, the smaller value within $\left| m_{ap} - m_{mo} \right|$ reveals the more commonalities between branches since they result in similar attention activations. 
Next, we conduct an element-wise minimize operation (Min) between $m_{ap}$ and $m_{mo}$ and perform a further Min-Max normalization (Norm) over the spatial dimension: 
\begin{equation}
    m_{co} = \text{Norm}(\text{Min}(m_{ap}, m_{mo})),
\end{equation}
where the output attention map $m_{co}$ can represent the common concerns of different branches since the larger value within $m_{co}$ means the smaller value within $\left| m_{ap} - m_{mo} \right|$. 

Similarly, we define the different concerns of branches by $m_{di}$, e.g., $1 - m_{co}$. 
Finally, the initial features of the common branch (the commonalities between branches) can be formulated as: 
\begin{equation}
    f^{'}_{co} = \frac{f_{ap} + f_{mo}}{2} * m_{co},
\label{eq:m_co}
\end{equation}
meanwhile, we refine the appearance and motion branch (highlight differences of each branch) by: 
\begin{equation}
    \begin{aligned}
        f^{'}_{ap} &= f_{ap} * m_{ap} * m_{di}, \\
        f^{'}_{mo} &= f_{mo} * m_{mo} * m_{di}
    \end{aligned}
\label{eq:m_di}
\end{equation}
as shown in Figure~\ref{fig:MultiGait++} (b).

\begin{table}[t]
\centering
\begin{tabular}{ccccc}
\hline
DataSet   & Batch Size & Milestones          & Steps \\ \hline
SUSTech1K & (8, 8, 10)     & (20k, 30k, 40k)         & 50k         \\
CCPG      & (8, 16, 30)    & (20k, 40k, 50k)         & 60k         \\ \hline
\end{tabular}
\caption{Implementation details. The batch size (q, p, k) indicates q IDs, p sequences per ID, and k frames per sequences.}
\label{tab:dataset}
\end{table}

\section{Experiment}
\subsection{Dataset} 
We conduct main experiments on the popular SUSTech1K and CCPG datasets, which offer publicly available RGB videos, allowing for flexible extraction of multiple gait modalities. 
Specifically, we utilize the officially provided silhouettes and personally perform the tasks of human parsing and optical flow extraction. 
Detailed pretreatment procedures are provided in the \textbf{Supplementary Materials}. 
The SUSTech1K offers many scenarios, including the normal(NM), bags(BG), clothing(CL), carrying(CR), umbrella(UB), uniform (UM), occlusion(OC), and nighttime(NT) conditions. 
Alternatively, the CCPG is designed around the challenges of clothing factors, featuring a diverse collection of full-(CL), up-(UP), down-clothing(DN), and bag(BG) changes. 
Our experiments strictly follow official evaluation protocols and take rank-1 accuracy(R-1) and mean average precision(mAP) as the primary metric.

GREW~\cite{zhu2021gait} and Gait3D~\cite{zheng2022gait3d} are two large real-world datasets. 
In recent years, more and more research~\cite{fan2022opengait,fan2024opengait,shen2024comprehensive} suggest paying more attention to real-world datasets.
Therefore, we conduct additional experiments on these two challenging real-world datasets, despite the absence of parsing and flow images in each dataset.

\begin{table*}[!t]
\centering
\setlength{\tabcolsep}{1mm}
\begin{threeparttable}
\begin{tabular}{c|c|cccccccc|cc}
\toprule
\multirow{2}{*}{Modality}          & \multirow{2}{*}{Method} & \multicolumn{8}{c|}{Probe Sequence (R-1)}                                                                     & \multicolumn{2}{c}{Overall}   \\
                                   &                                           & NM            & BG        & CL     & CR            & UB            & UN            & OC            & NT            & R-1            & R-5            \\ \midrule
\multirow{2}{*}{Skeleton}          & GaitGraph2~\cite{teepe2022towards}        & 22.2          & 18.2      & 6.8     & 18.6          & 13.4          & 19.2          & 27.3          & 16.4          & 18.6          & 40.2          \\
                                   & SkeletonGait~\cite{fan2023skeletongait}   & 55.0          & 51.0      & 24.7     & 49.9          & 42.3          & 52.0          & 62.8          & 43.9          & 50.1          & 72.6          \\ \midrule
\multirow{5}{*}{Sils}              & GaitSet~\cite{Chao2019}                   & 69.1          & 68.2      & 37.4     & 65.0          & 63.1          & 61.0          & 67.2          & 23.0          & 65.0          & 84.8          \\
                                   & GaitPart~\cite{fan2020gaitpart}           & 62.2          & 62.8      & 33.1     & 59.5          & 57.2          & 54.8          & 57.2          & 21.7          & 59.2          & 80.8          \\
                                   & GaitGL~\cite{gaitgl}                      & 67.1          & 66.2      & 35.9     & 63.3          & 61.6          & 58.1          & 66.6          & 17.9          & 63.1          & 82.8          \\
                                   & GaitBase~\cite{fan2022opengait}           & 81.5          & 77.5      & 49.6     & 75.8          & 75.5          & 76.7          & 81.4          & 25.9          & 76.1          & 89.4          \\
                                   & DeepGaitV2~\cite{fan2023exploring}        & 86.5          & 82.8      & 49.2     & 80.4          & 83.3          & 81.9          & 86.0          & 28.0          & 80.9          & 91.9          \\ \midrule
\multirow{2}{*}{Parsing}           & GaitBase$^{p}$~\cite{fan2022opengait}     & 80.3          &71.1       & 32.5     &72.6           &66.3           & 73.5        &81.7         &40.7               & 71.7          & 88.4              \\
                                   & DeepGaitV2$^{p}$~\cite{fan2023exploring}  & 84.7          & 77.3      & 29.4     & 78.2          & 75.8          & 80.2          & 87.9          & 43.3          & 77.3          & 91.3          \\ \midrule
Flow                               & DeepGaitV2$^{f}$~\cite{fan2023exploring}  & 84.4          & 82.8      & \textbf{54.9}     & 79.3          & 82.4          & 79.8          & 89.7          & 35.0          & 80.5          & 92.5          \\ \midrule
\multirow{2}{*}{Sils+Skeleton}     & BiFusion~\cite{peng2024learning}          & 69.8          & 62.3      & 45.4     & 60.9          & 54.3          & 63.5          & 77.8          & 33.7          & 62.1          & 83.4          \\
                                   & SkeletonGait++~\cite{fan2023skeletongait} & 85.1          & 82.9      & 46.6     & 81.9          & 80.8          & 82.5          & 86.2          & \textbf{47.5}   & 81.3          & 95.5          \\ \midrule
\multirow{2}{*}{Sils+Parsing+Flow} & MultiGait$^{s+p+f}$ (Ours)                 & 90.0          & 87.8      & 44.2     & 86.2          & 88.3       & 87.7          & 91.8          & 44.8      & 86.0          & 95.1          \\
                                   & MultiGait++ (Ours)         & \textbf{92.0} & \textbf{89.4} & 50.4 & \textbf{87.6} & \textbf{89.7} & \textbf{89.1} & \textbf{93.4} & 45.1 & \textbf{87.4} & \textbf{95.6} \\ \bottomrule
\end{tabular}
\end{threeparttable}
\caption{Evaluation with different attributes on SUSTech1K.}
\label{tab:multimodal-sustech1k}

\end{table*}
\begin{table*}[!t]

\centering
\setlength{\tabcolsep}{1mm}
\begin{threeparttable}
\begin{tabular}{c|cc|cc|cc|cc|cc|cl}
\toprule
\multirow{2}{*}{Modality}          & \multicolumn{2}{c|}{\multirow{2}{*}{Method}}                                                          & \multicolumn{2}{c|}{CL}        & \multicolumn{2}{c|}{UP}        & \multicolumn{2}{c|}{DN}        & \multicolumn{2}{c|}{BG}        & \multicolumn{1}{c}{Mean}      \\
\multicolumn{1}{c|}{}              & \multicolumn{1}{c}{}                                         &                       & \multicolumn{1}{c}{R-1} & mAP  & \multicolumn{1}{c}{R-1} & mAP  & \multicolumn{1}{c}{R-1} & mAP  & \multicolumn{1}{c}{R-1} & mAP  & \multicolumn{1}{c}{R-1}\\ \midrule
\multirow{3}{*}{Skeleton}                           & GaitGrapgh2~\cite{teepe2022towards}    & \multicolumn{1}{c|}{} & 5.0              &   2.5   & 5.3             &   4.0   & 5.8               &  4.2    & 6.2                     &  4.6    & 5.6                     &     \\
                                   & \multicolumn{2}{c|}{GaitTR~\cite{zhang2023spatial}}            & 15.7             & 9.7    & 18.3                   &   16.1   & 18.5         &  16.4    & 17.5              &   15.3   & 17.5                    &     \\
                                   & \multicolumn{2}{c|}{SkeletonGait~\cite{fan2023skeletongait}}   & 40.4             & 20.8    & 48.5                 &  35.8    & 53.0                    &   40.3   & 61.7                    &  48.1    & 50.9                    &     \\ \midrule
\multirow{4}{*}{Sils}              & \multicolumn{2}{c|}{GaitSet~\cite{Chao2019}}                   & 60.2             & 47.4 & 65.2                    & 63.2 & 65,1                    & 61.9 & 68.5                    & 65.8 & 64.8                    &     \\
                                   & \multicolumn{2}{c|}{GaitPart~\cite{fan2020gaitpart}}           & 64.3             & 48.0 & 67.8                    & 63.2 & 68.6                    & 63.8 & 71.7                    & 66.4 & 68.1                    &     \\
                                   & \multicolumn{2}{c|}{GaitBase~\cite{fan2022opengait}}           & 71.6             &  58.5    & 75.0                 &  71.6    & 76.8              &   73.3   & 78.6              &   75.3   & 75.5                    &     \\
                                   & \multicolumn{2}{c|}{DeepGaitV2~\cite{fan2023exploring}}        & 78.6             &  62.1    & 84.8                 &  81.5    & 80.7                    &  78.1    & 89.2                    &   85.8   & 83.3                    &     \\ \midrule
\multirow{2}{*}{Parsing}           & \multicolumn{2}{c|}{GaitBase$^{p}$~\cite{fan2022opengait}}     & 59.1             & 44.1     & 62.1            & 57.5     & 66.8                    &  61.9    & 68.1                    &  64.2    & 64.0                    &     \\
                                   & \multicolumn{2}{c|}{DeepGaitV2$^{p}$~\cite{fan2023exploring}}  & 69.6             &  51.7    & 75.8            &  71.4    & 75.8                    &   71.6   & 83.3                    &  79.3    & 76.1                    &     \\ \midrule
Flow                               & \multicolumn{2}{c|}{DeepGaitV2$^{f}$~\cite{fan2023exploring}}  & 68.8         &  49.5    & 73.3                    &  67.5    & 75.0                    &  68.0    & 76.8                    & 71.1     & 73.5                    &     \\ \midrule
\multirow{2}{*}{RGB}               & \multicolumn{2}{c|}{GaitEdge~\cite{liang2022gaitedge}}     & 66.9             & -     & 74.0            & -     & 70.6                    &  -    & 77.1                    &  -    & 72.2                    &     \\
                                   & \multicolumn{2}{c|}{BigGait~\cite{ye2024biggait}}  & 82.6             &  -    & 85.9            &  -    & \textbf{87.1}                    &   -   & \textbf{93.1}                    &  -    & 87.2                    &     \\ \midrule
\multirow{2}{*}{Sils+Skeleton}     & \multicolumn{2}{c|}{BiFusion~\cite{peng2024learning}}          & 62.6           &    46.7  & 67.6            &  64.1    & 66.3            &  61.9    & 66.0                    & 61.9     & 65.6                &     \\
                                   & \multicolumn{2}{c|}{SkeletonGait++~\cite{fan2023skeletongait}} & 79.1                    &   63.6   & 83.9            &   81.2   & 81.7                    & 79.3     & 89.9                    &   87.0   & 83.7                    &     \\ \midrule

\multirow{1}{*}{Sils+Parsing}     & \multicolumn{2}{c|}{XGait~\cite{zheng2024takes}}          & 72.8           &    59.5  & 77.0            &  74.3    & 79.1            &  75.7    & 80.5                    & 78.2     & 77.4                &       \\ \midrule
\multirow{2}{*}{Sils+Parsing+Flow} & \multicolumn{2}{c|}{MultiGait$^{s+p+f}$ (Ours)}                 & 81.3                    & 65.2 & 87.2                    & 83.6 & 82.9                    & 80.3 & 90.6                    & 87.4 & 85.5                    &     \\
                                   & \multicolumn{2}{c|}{MultiGait++ (Ours)}                         & \textbf{83.9}           & \textbf{68.5} & \textbf{89.0}   & \textbf{85.8} & 86.0    & \textbf{82.5} & 91.5     & \textbf{88.9} & \textbf{87.6}           &     \\ 

\bottomrule
\end{tabular}
\end{threeparttable}
\caption{Evaluation with different attributes on CCPG.}
\label{tab:multimodal-ccpg}

\end{table*}

\subsection{Implementation Details}
\label{sec:implementation_details}
1) Both the silhouette, human parsing, and optical flow images are aligned by the size-alignment method introduced by Takemura~\cite{Takemura2018} et al.
These images are further resized to 1×64×44, 1×64×44, and 3×64×44.
2) Table~\ref{tab:dataset} displays the main hyper-parameters of our experiments;
3) The spatial augmentation strategy suggested by OpenGait~\cite{fan2022opengait} is adopted;
4) The SGD optimizer with an initial learning rate of 0.1 and weight decay of 0.0005 is utilized.
5) A naive combination of MultiGait$^{s+p}$(input-level and cat fusion) and MultiGait$^{s+f}$(high-level and attention fusion), termed MultiGait$^{s+p+f}$ as shown in Figure~\ref{fig:MultiGait++} (c), is introduced to act as a strong baseline. 

\subsection{Comparsion around MultiGait++}
\noindent\textbf{Results on SUSTech1K: }
As shown in Table~\ref{tab:multimodal-sustech1k}, both MultiGait++ and its baseline, MultiGait$^{s+p+f}$, achieve rank-1 accuracies that significantly surpass other state-of-the-art methods in most cases.
This demonstrates the substantial benefits of multimodal modeling for gait recognition.
More importantly, MultiGait++ outperforms its baseline, demonstrating the effectiveness of our C$^2$Fusion strategy.

\begin{table*}[!t]
\centering
\begin{threeparttable}
\begin{tabular}{c|c|ccc|cccc}
\toprule
\multirow{2}{*}{Modality}          & \multirow{2}{*}{Method} & \multicolumn{3}{c|}{GREW}         & \multicolumn{4}{c}{Gait3D} \\ \cmidrule{3-9} 
                                   &                         & R-1   & R-5   & R-10  & R-1    & R-5   & mAP  & mINP \\ \midrule
\multirow{6}{*}{Sils}              & GaitSet~\cite{Chao2019} & 46.3 & 63.6 & 70.3 & 36.7  & 58.3 & 30.0 & 17.3 \\
                                   & GaitPart~\cite{fan2020gaitpart}                & 44.0 & 60.7 & 67.3 & 28.2  & 47.6 & 21.6 & 12.4 \\
                                   & GaitGL~\cite{gaitgl}                  & 47.3 &\multicolumn{2}{c|}{-}& 29.7  & 48.5 & 22.3 & 13.6 \\
                                   & GaitBase~\cite{fan2022opengait}                & 60.1 & 75.5 & 80.4 & 64.6  &      &  -   &      \\
                                   & QAGait~\cite{wang2024qagait}                  & 59.1 & 74.0 & 79.2 & 67.0  & 81.5 & 56.5 &  -   \\
                                   & DeepGaitV2~\cite{fan2023exploring}              & 77.7 & 88.9 & 91.8 & 74.4  & 88.0 & 65.8 &  39.2 \\ \midrule
\multirow{3}{*}{Skeleton}          & GaitGraph2~\cite{teepe2022towards}              & 33.5 &\multicolumn{2}{c|}{-}& 11.1 &      &  -   &      \\
                                   & GaitTR~\cite{zhang2023spatial}                  & 54.5 &\multicolumn{2}{c|}{-}& 6.6  &      &  -   &      \\
                                   & SkeletonGait~\cite{fan2023skeletongait}            & 77.4 & 87.9 & 91.0 & 38.1  & 56.7 & 28.9 & 16.1 \\ \midrule
\multirow{3}{*}{Sils+Skeleton}     & GaiRef~\cite{zhu2023gaitref}                 & 53.0 & 67.9 & 73.0 & 49.0  & 49.3 & 40.7 & 25.3 \\
                                   & MSAFF~\cite{zou2024multi}                   & 57.4 & 73.0 & 78.3 & 48.1  & 66.6 & 38.5 & 23.5 \\
                                   & SkeletonGait++~\cite{fan2023skeletongait}          & 85.8 & 92.6 & 94.3 & 77.6  & 89.4 & 70.3 & 42.6 \\ \midrule
\multirow{3}{*}{Sils+Flow}         & GaitFusion~\cite{feng2023fusion}              & 83.1 & 91.3 & 93.6 &       & \multicolumn{2}{c}{-} &      \\
                                   & MultiGait$^{s+f}$ (Ours)               & 91.4 & 96.4 & 97.5 &       &  \multicolumn{2}{c}{-}  &  \\
                                   & MultiGait++$^{s+f}$ (Ours)             & \textbf{93.4} & \textbf{97.3} & \textbf{98.3} &       &  \multicolumn{2}{c}{-}  &  \\ \midrule
\multirow{3}{*}{Sils+Parsing} & XGait~\cite{zheng2024takes}         &      & -    &    & 80.5  & 91.9 & 73.3 & 55.4 \\
                                   & MultiGait$^{s+p}$ (Ours)               &      & -    &    & 83.0  & 94.5 & 78.6 & 62.7 \\
                                   & MultiGait++$^{s+p}$ (Ours)             &      & -    &    & \textbf{85.4}  & \textbf{94.9} & \textbf{80.5} & \textbf{65.2} \\ 
\bottomrule
\end{tabular}
\end{threeparttable}
\caption{Recognition results on two real-world gait datasets, involving GREW, and Gait3D.}
\label{tab:real_world_dataset}
\end{table*}

\noindent\textbf{Results on CCPG: }
On the more challenging CCPG dataset, MultiGait++ outperforms all other SoTA methods across all conditions, as shown in Table~\ref{tab:multimodal-ccpg}. 
Excluding its baseline, MultiGait++ raises the SoTA standard by +4.8\%, +4.2\%, +4.3\%, and +1.6\% in rank-1 accuracy on the CL, UP, DN, and BG subsets, respectively. 
It also achieves considerable improvements compared to its baseline, i.e., +2.1\% in rank-1 accuracy averaged over CCPG. 
These results underscore the exceptional capability and practicality of MultiGait++ in handling complex clothing variations.

\noindent\textbf{More Results on Other Real-world Datasets: }
To further validate the effectiveness of MultiGait++, we conduct additional experiments on two challenging real-world datasets, GREW and Gait3D, despite the absence of parsing and flow images in each dataset. 
In this phase, we modify MultiGait++ (and MultiGait) into a two-branch input design, with a combination of silhouette and parsing for Gait3D, and silhouette and flow for GREW.

The results shown in Table~\ref{tab:real_world_dataset} demonstrate the notable superiority of both MultiGait and MultiGait++. 
Excluding its baseline MultiGait, MultiGait++ raises the SoTA standard by +7.6\% and +4.9\% in rank-1 accuracy on GREW and Gait3D. 
It is also worth mentioning that MultiGait++ continues to outperform its high-performance baseline, MultiGait, underscoring the robustness and effectiveness of the proposed C$^2$Fusion. 

\begin{table}[!t]
\centering
\setlength{\tabcolsep}{1mm} 
\begin{tabular}{c|cc|ccccc|c}
\toprule
\multirow{2}{*}{Index} &\multirow{2}{*}{\begin{tabular}[c]{@{}c@{}}$m_{co}$ \\ in Eq.\ref{eq:m_co}\end{tabular}} & \multirow{2}{*}{\begin{tabular}[c]{@{}c@{}}$m_{di}$ \\ in Eq.\ref{eq:m_di}\end{tabular}} & \multirow{2}{*}{NM} & \multirow{2}{*}{UB} & \multirow{2}{*}{UN} & \multirow{2}{*}{OC} & \multirow{2}{*}{NT} & \multirow{2}{*}{Overall} \\
    &                &                  &               &               &               &               &                &               \\ \midrule
(a) &$\times$        &    $\times$      & 90.0          & 87.8          & 87.7          & 91.8          &  44.8          & 86.0          \\
(b) &$\times$        &    \checkmark    & 90.1          & 88.7          & 87.9          & 93.0          &  \textbf{45.4} & 86.5          \\
(c) &\checkmark      &    $\times$      & 90.6          & 89.0          & 88.1          & 92.5          &  43.5          & 86.4          \\ 
(d) &\checkmark      &    \checkmark    & \textbf{92.0} & \textbf{89.7} & \textbf{89.1} & \textbf{93.4} &  45.1          & \textbf{87.4} \\ \bottomrule
\end{tabular}
\caption{
Ablation study on common concerns $m_{co}$ and different concerns $m_{di}$.
}
\label{tab:ablation}
\end{table}

\subsection{Ablation Study}
To validate the C$^2$ module's effectiveness, we conduct ablation experiments on $m_{co}$ and $m_{di}$ in Table~\ref{tab:ablation}.
Specifically, Table~\ref{tab:ablation} (b) means that removing $m_{co}$ but remaining $m_{di}$ in Figure~\ref{fig:MultiGait++} (b); 
Table~\ref{tab:ablation} (c)  means that removing the term of $m_{di}$ but remaining the term of $m_{co}$ in Figure~\ref{fig:MultiGait++} (b); 
Table~\ref{tab:ablation} (a) and~\ref{tab:ablation} (d) means the MultiGait$^{s+p+f}$ and MultiGait++.

Fusing these three modalities in various ways consistently yields high performance on SUSTech1K (compared to other SoTA methods in Table~\ref{tab:multimodal-sustech1k}), highlighting the benefits of multimodal methods. 
Furthermore, the C$^2$ module in MultiGait++, which preserves shared characteristics (by $m_{co}$) while emphasizing unique features (by $m_{di}$), further enhances recognition accuracy.

\section{Conclusion}
This work studies three typical gait modalities: silhouettes, human parsing, and optical flow. 
It emphasizes the importance of multimodal methods in gait recognition.
Furthermore, the proposed C$^2$Fusion strategy encourages each modality to highlight its unique features while preserving shared characteristics, effectively enriching the features for multimodal gait description. 
The integrated approach achieves superior accuracy and demonstrates that multimodal gait recognition has much to explore in the future. 

\section*{Acknowledgements}
This work was supported by the National Natural Science Foundation of China under Grant 62476120, and the Shenzhen International Research Cooperation Project under Grant GJHZ20220913142611021. 

\appendix

\section{Supplementary Material}
\begin{figure}[ht]
\centering
\includegraphics[width=1\linewidth]{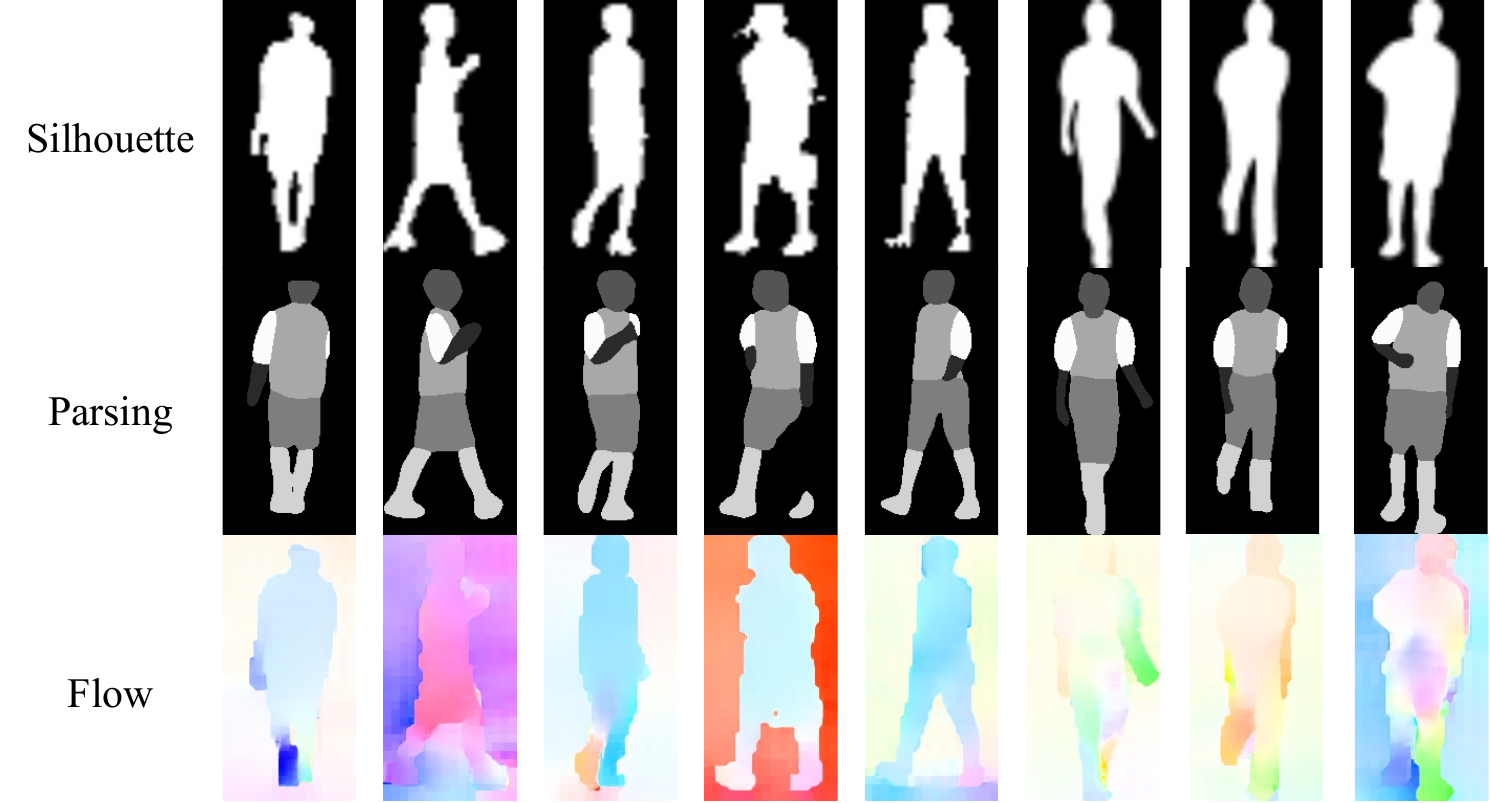}
\caption{Visualization of silhouette, human parsing, and flow}
\label{fig:intro}
\end{figure}
\subsection{Extraction of Human Parsing and Optical Flow}

Due to the unavailability of human parsing and flow data, we extracted these datasets ourselves. In this section, we will detail our extraction methods for both types of data, as shown in Figure~\ref{fig:intro}.

\noindent\textbf{Human parsing:} 
In recent years, researchers have developed the field of human parsing by investigating the structure of the human body and clothing components. QANet~\cite{yang2022quality}, proposed by Yang, has shown good performance on the Crowd Instance-level Human Parsing (CIHP)~\cite{gong2018instance} and Pascal-Person-Part~\cite{xia2017joint} datasets. The Pascal-Person-Part dataset divides the human body into six parts based on body part information, meeting the requirements for gait representation. To demonstrate the advantages of human parsing, we conducted further experiments using the pretrained model of QANet on the Pascal-Person-Part dataset to extract human parsing data.

\noindent\textbf{Optical flow:}
With the development of deep learning and the needs of downstream tasks, i.e., action recognition and video analysis, optical flow models designed for multi-frame scenarios have evolved continuously. VideoFlow~\cite{shi2023videoflow}, introduced in 2023 by Shi, achieved leading performance on the KITTI-2015\cite{menze2015object} and Sintel\cite{butler2012naturalistic} datasets. Consequently, we decided to use VideoFlow to generate corresponding optical flow images.
The RGB images released by CCPG and SUSTech1K are not the original RGB images but images extracted from bounding boxes. 
RGB images are first padded and resized to make their sizes consistent, ensuring that the proportions of human bodies in the images remain consistent.
Optical flow captures motion information by analyzing pixel changes between consecutive frames. Gait features focus on the human body. However, padding and resize operations directly on RGB images may introduce background trajectory noise in the generated optical flow images, deviating from the requirements for gait features. Therefore, silhouettes are used to mask non-human structures in the RGB images after padding and resizing, and the masked RGB images are then used to obtain the optical flow images.

\label{sec:dataset}
\begin{figure}[htbp]
\centering
\includegraphics[width=0.48\textwidth]{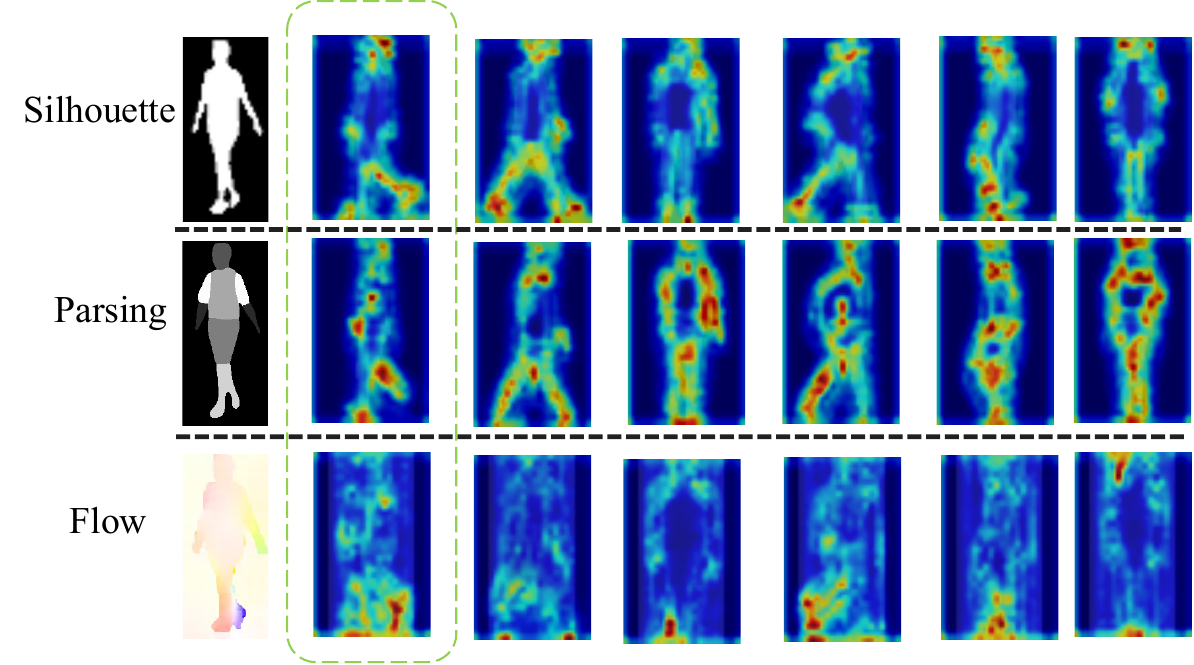}
\caption{The heatmaps ~\cite{zhou2016learning} of MultiGait$^s$ v.s. MultiGait$^p$ and MultiGait$^f$. Each row corresponds to the same modality, while each column is sourced from the same RGB image.}
\label{fig:heatmap}
\end{figure}
\subsection{More Experimental Results}
\noindent\textbf{Silhouette, human parsing, and flow. }
The image displays three different techniques for gait: Silhouette, Parsing, and Flow, using heatmaps, as shown in Figure~\ref{fig:heatmap}.

\begin{enumerate}
    \item \textbf{Silhouette}:
    The Silhouette focuses on the overall shape of the human body. The heatmap shows the body's outline, with hot spots mainly in the center and ends of the limbs, capturing the general posture and shape.
    \item \textbf{Human parsing}:
    The heatmap for Parsing focuses on the human body's outline and internal structural information.
    \item \textbf{Flow}:
    Flow emphasizes the parts of the body in motion. The varying intensity of colors in the heatmap highlights the most active areas during movement, like the legs and arms, which are useful for capturing dynamic motion and analyzing movement patterns.
\end{enumerate}

Heatmaps help us better understand the characteristics of each modality and their complementary aspects, providing a comprehensive view of human dynamics and structure.

\noindent\textbf{Parameter and GFLOPs of MultiGait++. }
Real gait recognition applications require several essential steps such as detection, segmentation, and final recognition. As shown in Table~\ref{tab:gflops}, the multimodal MultiGait++ introduces only a modest increase in computation cost, about 10\% to 15\% as listed below, rather than a large multiple, to achieve significant accuracy gains of +4.3\% on CCPG, +6.9\% on SUSTech1K, +15.7\% on GREW, and +11.0\% on Gait3D, compared to the strong DeepGaitV2.
\begin{table*}[!t]
\centering
\begin{tabular}{c|cccc|c}
\toprule
Index & Step                                     & Method                                                     & Backbone                   & Resolution                                                & GFLOPs    \\ \midrule
1     & Detection                                & Yolov5-x                                                   & CSPDarknet53               & 1024$^2$                                                  & 316.9     \\
2     & Segmentation                             & DeepLabV3                                                  & ResNet50                   & 256$^2$                                                   & 28.5      \\
3     & Parsing                                  & SCHP                                                       & A-CE2P                     & 256$^2$                                                   & 21.9      \\
4     & Flow                                     & VideoFlow                                                  & ResNet-like                & 256$^2$                                                   & 21.0      \\
5     & Gait Recognition (GR)                    & DeepGaitV2                                                 & ResNet-like                & 64$\times$44                                              & 0.8       \\
6     & Unimodal GR: 1+2+5$\times$1              & DeepGaitV2                                                 & ResNet-like                & -                                                         & 346.2     \\
7     & Multimodal GR: 1+2+3+4+5$\times$3        & MultiGait++                                                & ResNet-like                & -                                                         & 390.7     \\ \bottomrule
\end{tabular}
\caption{Parameter and GFLOPs of the gait recognition pipeline. Note: Techniques such as weight pruning can effectively reduce the computation cost of various CNN-based models. Here we show the unreduced ones to ensure a fair comparison across different steps.}
\label{tab:gflops}
\end{table*}

\begin{table*}[!hbt]
\centering
\begin{tabular}{c|c|cccccccc|c}
\toprule
Method      & SkeletonMap & NM            & BG            & CL            & CR            & UB            & UN            & OC            & NT            & R-1           \\ \midrule
MultiGait++ &  $\times$   & \textbf{92.0} & \textbf{89.4} & 50.4          & 87.6          & \textbf{89.7} & \textbf{89.1} & 93.4          & 45.1          & 87.4          \\
MultiGait++ & \checkmark  & 90.5          & 89.3          & \textbf{53.1} & \textbf{88.5} & 88.6          & 89.0          & \textbf{93.5} & \textbf{57.0} & \textbf{87.8} \\ \bottomrule
\end{tabular}
\caption{Effect of Skeleton Map on MultiGait++ on SUSTech1K.}
\label{tab:moremodality}
\end{table*}




\noindent\textbf{Adding more of MultiGait++. }
Adding gait modality to Table~\ref{tab:fusion} would significantly increase the number of experiments needed for a fair comparison. So, we chose to directly concatenate the skeleton map~\cite{fan2023skeletongait} with the silhouette and parsing at the input level to enhance MultiGait++. The results in Table~\ref{tab:moremodality} show that incorporating the skeleton in this simple manner has limited impact on improving MultiGait++.

The reason may be that the parsing already provides some structural information. Alternatively, a more sophisticated integration of the skeleton within MultiGait++ could yield better results. Overall, we hope this work encourages further exploration of diverse gait modalities, including skeleton, depth images, SMPL, and more.

\bibliography{aaai25}
\end{document}